\documentclass[letterpaper, 10 pt, conference]{ieeeconf}  % Comment this line out if you need a4paper

\IEEEoverridecommandlockouts                              % This command is only needed if 
\usepackage{graphics} % for pdf, bitmapped graphics files
\usepackage{epsfig} % for postscript graphics files
\usepackage{times} % assumes new font selection scheme installed
\usepackage{amsmath} % assumes amsmath package installed
\usepackage{amssymb}  % assumes amsmath package installed
\usepackage{algorithm}
\usepackage{algpseudocode} 
\usepackage{cite}
\newtheorem{lemma}{Lemma}
\newtheorem{theorem}{Theorem}
\newtheorem{definition}{Definition}
\newtheorem{proposition}{Proposition}
\newtheorem{corollary}{Corollary}
\newtheorem{remark}{Remark}
\usepackage{subfigure}
\usepackage{xcolor}
\usepackage{hyperref}
\newcommand{\fl}{{(i)}}
\title{\LARGE \bf
Federated Nonlinear System Identification
}

\author{Omkar Tupe$^{*}$, Max Hartman$^{*}$, 
Lav R. Varshney, and Saurav Prakash % stops a space
\thanks{Omkar Tupe is with the Wadhwani School of Data Science and AI, and Saurav Prakash is with the Department of Electrical Engineering and the Center for Responsible AI, Indian Institute of Technology Madras. Max Hartman is with the Department of Electrical and Computer Engineering and the Coordinated Science Laboratory, University of Illinois Urbana-Champaign. Lav R. Varshney is with the AI Innovation Institute, Stony Brook University.
Email: \texttt{omkar.tupe@dsai.iitm.ac.in, saurav@ee.iitm.ac.in, maxh3@illinois.edu, lav.varshney@stonybrook.edu }}}
\begin{document}

\maketitle
\thispagestyle{empty}
\pagestyle{empty}

%%%%%%%%%%%%%%%%%%%%%%%%%%%%%%%%%%%%%%%%%%%%%%%%%%%%%%%%%%%%%%%%%%%%%%%%%%%%%%%%
\begin{abstract}

We consider federated learning of linearly-parameterized nonlinear systems. We establish theoretical guarantees on the effectiveness of federated nonlinear system identification compared to centralized approaches, demonstrating that the convergence rate improves as the number of clients increases. Although the convergence rates in the linear and nonlinear cases differ only by a constant, this constant depends on the feature map $\phi$, which can be carefully chosen in the nonlinear setting to increase excitation and improve performance. We experimentally validate our theory in physical settings where client devices are driven by i.i.d.\ control inputs and control policies exhibiting i.i.d.\ random perturbations, ensuring non-active exploration. Experiments use trajectories from nonlinear dynamical systems characterized by real-analytic feature functions, including polynomial and trigonometric components, representative of physical systems including pendulum and quadrotor dynamics. We analyze the convergence behavior of the proposed method under varying noise levels and data distributions. Results show that federated learning consistently improves convergence of any individual client as the number of participating clients increases. 
\let\thefootnote\relax\footnotetext{* denotes equal contribution}
\end{abstract}
% {\color{red} For Lav: Please add the funding support statement in Acknowledgement}
\section{Introduction}
Dynamical system models describe how a system evolves over time based on its current state, control inputs, and external disturbances. These models are important to fields such as control theory, physics, and robotics \cite{willems1989models}. A common category of systems is linear time-invariant (LTI) systems, where both the next state and output are linear functions of the current state and input. These systems are governed by parameters that remain constant over time \cite{SibaiH2018}. In contrast, nonlinear functions determine these relationships in nonlinear systems~\cite{guckenheimer2013nonlinear, Jiang_2019}.

\begin{figure}[!t]
\centering
\includegraphics[width=0.8\linewidth]{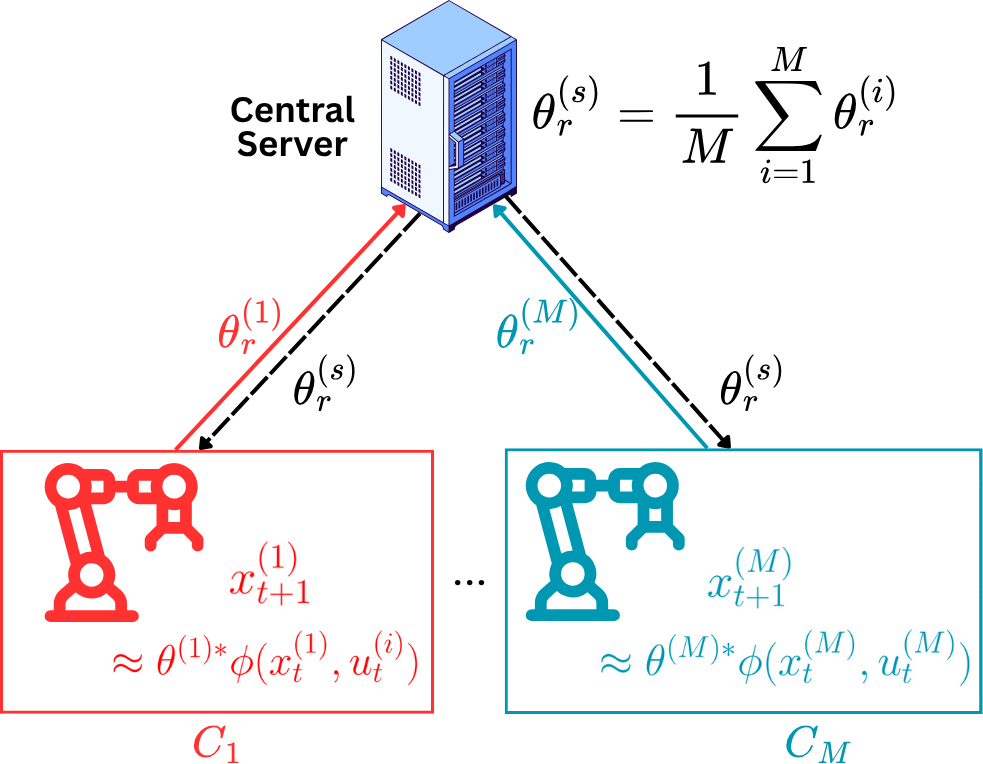}
     \caption{Federated learning framework for nonlinear dynamical system identification involving $M$ clients which are similar but non-identical in nature  and a central server. In each global communication round $r$, client $C_i$ receives the global model $\theta_s$ performs local updates using its own trajectories data, and transmits the locally updated model $\theta_i$ back to the server. The server then aggregates the models to obtain an updated global model for the next round.
     }
\label{fig:Nonlinear_FL}
\end{figure}

System identification is a data-driven approach to learn dynamical models from input and output data. It is often used to learn dynamical systems for which exact dynamics are unknown, with applications including robotics, fluid dynamics, physics and healthcare \cite{sarkar2021finite, venkatesh2002system,NAP26894}. Traditional approaches for system identification assume centralized data. However, in modern applications, data is often distributed across many devices, and transmitting all data to a central server is often impractical due to privacy, bandwidth, or energy limitations. To address these challenges, we consider \textit{federated} system identification, which is emerging as a privacy-preserving alternative. In this paradigm, multiple clients collaboratively learn a model without sharing raw data, communicating only model updates. Prior work has explored federated identification for LTI systems \cite{fedsysid}.

In this paper, we study the problem of federated nonlinear system identification. We consider a setting where $M$ distributed clients each observe trajectory data from different nonlinear dynamical systems that belong to the same underlying family (see Figure \ref{fig:Nonlinear_FL}). The dynamical systems we consider exhibit heterogeneity, modeled by the heterogeneity parameter $\epsilon$. Federated learning should be used for system identification when multiple similar systems share common structure and data cannot be centralized, enabling privacy-preserving collaborative learning. Results show that aggregating across more participating clients improves convergence and reduces individual client noise.

% The goal of this approach is to identify a shared nonlinear model that generalizes across these clients while respecting local variability and preserving data privacy.
% \begin{figure*}[t]
% \centering
% \subfigure{
%     \begin{minipage}[t]{0.28\textwidth}
%     \centering
%     \includegraphics[width=\textwidth,trim=0 0 0 30, clip]{Pendulum_sys_id_eps_0_01.png}
%     \vspace{1mm}
%     % \textbf{[a]}
%     (a)
%     \label{fig:Pend_4_M}
%     \end{minipage}
% }
% \hfill
% \subfigure{
%     \begin{minipage}[t]{0.28\textwidth}
%     \centering
%     \includegraphics[width=\textwidth,trim=0 0 0 30, clip]{Pendulum_sys_id_eps_5.png}
%     \vspace{1mm}
%     % \textbf{[b]}
%     (b)
%     \label{fig:Pend_4_N}
%     \end{minipage}
% }
% \hfill
% \subfigure{
%     \begin{minipage}[t]{0.28\textwidth}
%     \centering
%     \includegraphics[width=\textwidth,trim=0 0 0 30, clip]{pend_error_vs_clients_count_var_eps.png}
%     \vspace{1mm}
%     % \textbf{[c]}
%     (c)
%     \label{fig:Pend_4_eps}
%     \end{minipage}
% }
% \caption{Impact of client count and heterogeneity on  federated nonlinear system identification. (a) Estimation error versus number of clients $M$ at fixed trajectories per client $N_i = 10$ and heterogeneity bound $\epsilon = 0.01$; (b) Estimation error versus $M$ for the same $N_i$ with $\epsilon = 5$, illustrating increased model‐mismatch tolerance; (c) Estimation error as a function of $\epsilon$ for fixed $M$ and $N_i$, showing sensitivity of global model accuracy to client‐level parameter variations.}
% \label{fig:pend_sy_id_m_eps}
% \end{figure*}
% \subsection{Key Contributions}
\vspace{1pt}
Our main contributions are as follows:
\begin{itemize}
    \item \textbf{Nonlinear framework.} We consider linearly-parameterized nonlinear dynamical systems, with a focus on the widely used piecewise affine (PWA) model for representing nonlinear dynamics \cite{mania2022active, musavi2024identification}. In PWA models, state transitions are expressed as linear functions of nonlinear feature embeddings of state-input pairs. While prior work has explored federated system identification of linear dynamical systems, to the best of our knowledge, this work develops the first federated learning framework for system identification of PWA-based nonlinear dynamical systems.
    \item \textbf{Convergence analysis.} We analyze the convergence of the federated nonlinear system identification problem. Our results show that clients benefit significantly by collaborating through federated learning. Particularly, we show that the convergence error decreases as $\tilde{\mathcal{O}}(1/\sqrt{M})$, where $M$ is the number of clients and  $\tilde{\mathcal{O}}(\cdot)$ is the soft-O notation, thus having significant improvement in convergence as more clients collaborate. Our results also theoretically characterize how heterogeneity across clients impacts the convergence error.  
    
    \item \textbf{Experimental analysis.} 
    We conduct experiments analyzing the convergence behavior of our approach as a function of the total number of participating clients, the number of local samples at each client, as well the heterogeneity of dynamical systems across clients. Complete details are provided in Section~\ref{sec:experiments}. 
    \item \textbf{Algorithm validation.} Our experiments demonstrate that our federated approach achieves an improved convergence rate compared to a single-client system.
\end{itemize}

\section{Related Works}
\subsection{System Identification} The paper \cite{ljung1998system} provides an overview of mathematical methods for building models of dynamical systems using input and output signal measurements, focusing on how parameter estimation techniques can be applied to characterize system behavior in time and frequency domains. This approach is instrumental in transforming observed data into predictive models, supporting applications in signal processing and control. System identification can be extended to linearly-parameterized nonlinear systems~\cite{musavi2024identification}. The authors in \cite{JedP20}  present a robust finite-time error bound for identifying stable linear systems using the ordinary least-squares (OLS) estimator. The study in \cite{pmlr-v75-simchowitz18a} also analyzes OLS identification from a single trajectory for stable and marginally stable linear time-invariant systems, using the block-martingale small-ball method to address mixing dependencies and establishes nearly optimal estimator rates for a wide class of linear time series. The work further provides tight, regime-specific finite-time error bounds for OLS identification of general linear dynamical systems. The framework in \cite{sarkar2019near} establishes that OLS is optimally consistent only under specific regularity conditions, unifying previous results across all spectral regimes.
This approach struggles with nonlinear or complex signal structures, often requiring significant computational resources and large datasets to achieve reliable identification.

\subsection{Federated Learning}
Federated learning is a machine learning paradigm which allows clients to collaboratively train a model while preserving data privacy \cite{mcmahan2023communicationefficientlearningdeepnetworks}. Each client possesses a local dataset and sends weight updates to the server. The server then aggregates these weights and updates the global model. The method presented by \cite{truex19} is a hybrid federated learning approach that combines differential privacy and secure multi-party computation to enhance privacy protection while maintaining high model accuracy. Communication-efficient federated learning strategies through structured low-dimensional updates and compressed sketched updates may achieve a substantial reduction in communication overhead while preserving model accuracy \cite{kon16}. The work of\cite{Sahu2018OnTC} introduces the FedProx framework, which adds a proximal term to FedAvg and provides convergence guarantees to achieve more robust and stable training in the presence of statistical and system heterogeneity across clients. Federated learning  architectures can use a peer-to-peer approach \cite{yang2019federated}, however, our discussion focuses on the traditional server-client model, in which a central server periodically aggregates updates from multiple clients. Although most federated learning works are concerned with learning deep neural networks, our work focuses on understanding the theory and implementation of federated system identification of nonlinear dynamical systems.

\subsection{Federated System Identification} 
The goal of federated system identification is to learn a shared estimate of the system parameters that yields a small estimation error relative to the true parameters of each client $i$'s true system parameters $\theta^{(i)*}$. Clients $i$ do not share their private data with the server, and each client independently estimates its unknown system dynamics. The paper \cite{fedsysid} formulated an FL framework for system identification of linear system models across heterogeneous clients with improved sample efficiency, proposing the FedSysID algorithm and showing that more clients improve the model convergence rate. However, \cite{fedsysid} was restricted to synthetic linear systems. The authors of \cite{11077727} introduce Koopman learning as a method for dealing with nonlinear systems that have partially observable data. The setting is where multiple clients observe an identical nonlinear system, but the system lacks any active control ($u_t)$, which is contrary to what we will consider in \eqref{eq:system}.
These limitations motivate our work on federated nonlinear system identification.

\section{Federated Nonlinear System Identification}
In this section, we formally define the federated nonlinear system identification framework. Then we provide a convergence rate of the system with respect to the number of clients, number of trajectories at each client,  trajectory length, and heterogeneity across clients. Finally, we present the \texttt{FNSysId} algorithm that describes the client-server interactions for nonlinear system identification.
\subsection{Preliminaries}
The formal objective of the federated system identification problem is to estimate a model that accurately captures the collective behavior of the participating systems. Formally, first suppose that for each client $i\in[M]$, we have the following linearly-parameterized nonlinear dynamics:
\begin{equation}
x_{t+1}^{(i)}=\theta^{(i)*}\phi(x_t^{(i)}, u_t^{(i)})+w_t^{(i)}, \label{eq:system}
\end{equation}
where $x_t^{(i)}$, $u_t^{(i)}$, and $w_t^{(i)}$ denote the state, control, and disturbance respectively. Furthermore, $\phi: \mathbb{R}^{n_x}\times \mathbb{R}^{n_u}\to \mathbb{R}^{n_\phi}$ denotes a vector of known nonlinear mappings \cite{mania2022active,musavi2024identification}.

In the following, we recall two key definitions from \cite{musavi2024identification} that will be used in our assumptions. Refer to \cite{musavi2024identification} for further clarification on these definitions.
\begin{definition}
    (Semi-continuous distribution). A Euclidean probability distribution $\mathbb{P}$ is semi-continuous if there does not exist a set $E$ with Lebesgue measure zero such that $\mathbb{P}(E)=1$. 
\end{definition}
\begin{definition}
    (Local input-to-state stability). For the general nonlinear system $x_{t+1}=f(x_t,d_t)$ with $x_t\in \mathbb{R}^{n_x}$, $d_t\in\mathbb{R}^{n_d}$, let $f$ be a continuous function with $f(0,0)=0$. Then, the system is defined to be locally input-to-state stable (LISS) if there exist constants $\rho_x>0$, $\rho>0$ and functions $\gamma \in \mathcal{K}$, $\beta\in\mathcal{KL}$ such that for all $x_0\in\{x_0\in\mathbb{R}^{n_x}:||x_0||_2\leq \rho_x\}$ and any input $d_t\in\{d\in \mathbb{R}^{n_d}:\sup_t||d_t||_\infty\leq \rho \}$, it holds that $||x_t||_2\leq \beta(||x_0||_2, t)+\gamma(\sup_t||d_t||_\infty)$ for all $t\geq 0$.\footnote{We provide the definitions of K and KL in the Appendix.}
\end{definition}

We next provide four assumptions for the local dynamical systems at clients \cite{musavi2024identification}  (we omit client indexing for brevity).

\begin{itemize}
    \item Assumption 1: All components of the feature vector $\phi(\cdot)$ are real-analytic functions in $\mathbb{R}^{n_x+n_u}$, i.e., all components in $\phi(\cdot)$ are infinitely differentiable.
    \item Assumption 2: Noise $w_t$ is i.i.d., and  follows a semi-continuous, zero-mean distribution with positive semi-definite covariance matrix $\sum_{w}\succeq \sigma_w^2I_{n_x}$.
    \item Assumption 3: Input $u_t$ is i.i.d following a semi-continuous, zero-mean distribution with positive semi-definite covariance matrix $\sum_{u}\succeq \sigma_u^2I_{n_x}$.
    \item Assumption 4: The system \eqref{eq:system} is LISS with parameters $\rho_x$ and $\rho$ such that $\rho_x\geq||x_0||_2$ and $\rho\geq \max(w_{\max}, u_{\max})$.
\end{itemize}
Additionally, inspired by \cite{fedsysid}, we consider the following assumption of bounded system heterogeneity across clients. 
\begin{itemize}
\item Assumption 5: $\max_{i,j\in[M]}||\theta^{(i)*}-\theta^{(j)*}|| \leq \epsilon$.
\end{itemize}
 Our goal is to solve the least squares error estimate problem of our system to find the optimal system parameters:
\begin{align}\bar\theta_{LSE}=\frac{1}{M}\sum_{i=1}^M\arg\min_{\theta\in\mathbb{R}^{n_\phi \times n_x}}\bigl\|X_{+}^{(i)}- \theta\,\Phi^{(i)}\bigr\|_F^{2},
\end{align}
where $X_{+}^{(i)}=\bigl[x_0^{(i,1)},\dots, x_{T-1}^{(i, N_i)}\bigr]$, and $\Phi^{(i)}\in\mathbb{R}^{n_\phi\times(N_iT)}$ is defined as follows:
\begin{align*}
\Phi^{(i)}=\bigl[\phi(x^{(i,1)}_{0},u^{(i,1)}_{0}),\dots,\phi(x^{(i,N_i)}_{T-1},u^{(i,N_i)}_{T-1})].\nonumber
\end{align*}
Functionally, this optimization estimates each client's local dynamics via least-squares regression, then averages the results to obtain a global model.

\subsection{Convergence Rate}
A least-squares estimate is meaningful only if the design matrix built from each client’s regressor vectors is well conditioned.  In linear system identification this role is played by
a \emph{persistent‐excitation} assumption.  For linearly-parameterized nonlinear dynamics, we recover a similar property through the
block‑martingale small‑ball (BMSB) condition (Definition \ref{def:bmsb}). Lemma~\ref{lemma:1} is used to show that the analytic feature map and new inputs guarantee the required excitation. 

\begin{lemma}[BMSB for open‑loop systems]\label{lemma:1}
Let each client \(i\in[M]\) run open‑loop inputs \(u^{(i)}_t=\eta^{(i)}_t\). Let client $i$ collect a trajectory of length $T$. For every $t=0,1,\dots,T-1$, define the filtration
\[
\mathcal F^{(i)}_t
   :=\sigma\bigl(
        w^{(i)}_{0},\dots,w^{(i)}_{t-1},\;
        x^{(i)}_{0},\dots,x^{(i)}_{t},\;
        \eta^{(i)}_{0},\dots,\eta^{(i)}_{t}
     \bigr).
\]
Under Assumptions~1–4 there exist constants \(s_\phi>0\) and
\(p_\phi\in(0,1)\) (as defined by \cite{musavi2024identification}) such that, for every client \(i\) and every unit vector
\(v\in\mathbb S^{n_\phi-1}\),
\begin{align}
\Pr\!\bigl(
      |v^{\!\top}\phi(x^{(i)}_{t},u^{(i)}_{t})|
        \,\ge s_\phi
      \;\big|\;
      \mathcal F^{(i)}_{t-1}
    \bigr)
    \;\ge\; p_\phi
    \quad\text{a.s.}
\end{align}
Hence the regressor process \(\{\phi(x^{(i)}_{t},u^{(i)}_{t})\}_{t\ge0}\)
satisfies the \((1,s_\phi^{2}I_{n_\phi},p_\phi)\)-BMSB condition for every
client \(i\).
\end{lemma}
\begin{proof}
   Please refer to the Appendix for the proof.
\end{proof}

Each new regressor $\phi(x^{(i)}_{t},u^{(i)}_{t})$ has (conditioned
on the past) at least probability $p_\phi$ of having length $s_\phi$ in \emph{every}
direction, hence the excitation we desired.

To use least-squares, the empirical Gram matrices must be invertible. Proposition 1 provides that guarantee and will be used to determine
the \(1/N_{\mathrm{tot}}\) factor in our final error bound.
\begin{proposition} \label{prop:1} Fix $\delta\in (0,1)$. Suppose that each client $i$ collects $N_i$ samples of length $T$. For every client $i\in[M]$, where $N_{\mathrm{tot}}:=\sum_{i=1}^{M}N_iT$,
if each sample size satisfies $N_iT\;\ge\;
\frac{4}{p_\phi}[n_\phi\log (9)+\log(M/\delta)],$ then, with probability at least $1-\delta$,
\[
\lambda_{\min}\!\bigl(\Phi^{(i)}\Phi^{(i)\!\top}\bigr)
     \;\ge\;\tfrac12\,s_\phi^{2}N_iT,
     \quad i\in[M],
\]
and hence the pooled Gram matrix
\begin{align}
G:=\sum_{i=1}^{M}\Phi^{(i)}\Phi^{(i)\!\top}
      \succeq \tfrac12\,s_\phi^{2}N_{\mathrm{tot}}\,I_{n_\phi}.
\end{align}
\end{proposition}
\begin{proof}
   Please refer to the Appendix for the proof.
\end{proof}
We now need to bound the stochastic noise regressor cross-term. This will be used to guarantee that the estimator remains stable and accurate despite disturbances. 
\begin{proposition}\label{prop:2}
Fix $0<\delta<1$.  
For client $i\in[M]$ define the noise matrix
\[
W^{(i)}:=[w^{(i)}_{0}\;w^{(i)}_{1}\;\dots\;w^{(i)}_{N_i-1} ]\in\mathbb{R}^{n_x\times N_iT},
\]
and design matrix
\[
\Phi^{(i)} := \big[\phi(x^{(i,j)}_t,u^{(i,j)}_t)\big]_{\substack{j=1,\dots,N_i \\[2pt] t=0,\dots,T-1}}
  \in \mathbb{R}^{n_\phi \times N_i T}.
\]
Then, with probability at least $1-\delta$,
\begin{align}
\| W^{(i)}(\Phi^{(i)})^\top \|_2
   \le
   4\sigma_w
   \sqrt{\,N_iT(n_x+n_\phi+\log(2M/\delta))}
\end{align} for $i\in[M]$.

Consequently, for the pooled noise matrix
\(
P:=\sum_{i=1}^{M}W^{(i)}(\Phi^{(i)})^{\!\top},
\)
the same event implies
\begin{align}
\|P\|_2
   \;\le\;
   4\sigma_w
   \sqrt{\,N_iT(n_x+n_\phi+\log(2M/\delta))}.
\end{align}

\end{proposition}
\begin{proof}
   Please refer to the Appendix for the proof.
\end{proof}
These auxiliary propositions can now be combined to bound the final convergence rate.

\begin{theorem}[Finite‑sample error]\label{theorem:1}
Let $s_\phi$, $p_\phi$ be as in Lemma \ref{lemma:1} and set
\[
C_1:=\frac{8\sigma_w}{s_\phi^{2}},\qquad
\,C_2:=\frac{b_\phi}{s_\phi^{2}}+\frac{1}{2}.
\]
Then for any $\delta\in(0,1)$, with probability at least $1-3\delta$,
\[
\bigl\|\bar\theta_{\mathrm{LSE}}-\theta^{(i)\!*}\bigr\|_2
  \;\le\;
  C_1\sqrt{\frac{n_x+n_\phi+\log(2M/\delta)}{T\sum_{i=1}^MN_i}}
  + C_2\,\varepsilon
\] for all $i\in[M].$
\end{theorem}
\begin{proof}
    Please refer to the Appendix for the proof.
\end{proof}

\begin{remark}
The first term in the convergence error decreases as $1/\sqrt{T\sum_{i=1}^MN_i}$, where $\sum_{i=1}^MN_i$ is the total number of trajectories across all clients. Furthermore, since the number of trajectories $N_i$ at client $i$ is a constant for a given trajectory length $T$, the error decreases as  $\tilde{\mathcal{O}}(1/\sqrt{M})$.
% , where $\tilde{\mathcal{O}}(\cdot)$ denotes the soft-O notation. 
\end{remark}
\begin{remark}

For smaller $\epsilon$, the underlying system dynamics at the clients are more similar, enabling FL clients to effectively use data from all participants. This enhances the performance of each client, particularly achieving a convergence speedup of ${\approx}\sqrt{M}$.  
\end{remark}

\iffalse\red{Saurav: Add some remarks on theorem result. like error is $1/sqrt{NM}$, so if more clients participate, each client can have a smaller sample complexity. furthermore, for a fixed sample complexity, the error goes down as $1/sqrt{M}$. also, the convergence error bound has a constant error, dependent only on $\epsilon$. thus for large $\epsilon$, the gains due to increase in M may not be significant.}\fi
\subsection{FNSysId}
% \href{}{}
The \texttt{FNSysId} \footnote{The source code is available for download - \url{https://github.com/OMKARTT/FNSysId}} implementation is formalized in Algorithm \ref{FNSysId_algo}, which shows how clients perform computations and synchronize model updates with the central server during each communication round. This is the standard FL algorithm.
% The \texttt{FNSysId} implementation is formalized in Algorithm \ref{FNSysId_algo}, which shows how clients perform local computations and synchronize model updates with the central server during each communication round. This is the standard FL algorithm. 

% \begin{algorithm}
% \caption{\texttt{FNSysId}: Federated Nonlinear System Identification}
% \small
% \label{FNSysId_algo}
% \begin{algorithmic}[1]
% \State \textbf{Initialize} server with global parameter $\bar{\theta}_{0}$ and learning rate $\alpha$ ;
% \State \textbf{Initialize} each client $i \in [M]$ with local parameter $\theta^{(i)}_{0,0} = \bar{\theta}_{0}$ ;
% \For{global iteration $r = 0,1,\ldots,R-1$}
%     \State $\rhd$ \textbf{Client side:}
%     \For{each client $i \in [M]$ in parallel}
%         \State $\theta^{(i)}_{r+1} = \texttt{NonlinearClientUpdate}\big(i, \bar{\theta}_{r}, K_i\big)$ (\ref{eq:fedavg})
        
%         \State Send $\theta^{(i)}_{r+1}$ to server
%     \EndFor
%     \State $\rhd$ \textbf{Server side:}
%     \State $\bar{\theta}_{r+1} = \frac{1}{M} \sum_{i=1}^M \theta^{(i)}_{r+1}$
    
%     \State Broadcast $\bar{\theta}_{r+1}$ to all clients
% \EndFor
% \State \textbf{Return} global estimate $\bar{\theta}_{R}$
% \end{algorithmic}
% \end{algorithm}

Each client initializes a guess for its local model parameters, denoted as $\bar{\theta}_0$, and a learning rate $\alpha$. These local initializations are critical for algorithm convergence and allow each client to tailor its optimization process based on its individual system characteristics and data availability.

In our framework, we assume all clients $\mathcal{C} = \{C_1, C_2, \ldots, C_M\}$ actively participate in every global iteration~$r\in[R]$. 

Each client performs $K_i$ local updates independently using its own local data before any communication with the central server (line 6), where \texttt{NonlinearClientUpdate} iteratively executes:
\begin{align}
\theta_{r,k}^{\fl} =\; & \theta_{r,k-1}^{\fl} 
+ \alpha \left({y}^{\fl} - \theta_{r,k-1}^{\fl} {X}^{\fl}\right)(X^{(i)})^T,
\label{eq:fedavg}
\end{align}
where $X^{(i)}$ is the input data and $y^{(i)}$ is the true system. $k = 1,2,\ldots,K_i.$ Note that this is a restatement of \texttt{FedAvg}.
% \begin{algorithm}
% \caption{\texttt{FNSysId}: Federated Nonlinear System Identification}
% \small
% \label{FNSysId_algo}
% \begin{algorithmic}[1]
% \State \textbf{Initialize} server with global parameter $\bar{\theta}_{0}$ and learning rate $\alpha$ ;
% \State \textbf{Initialize} each client $i \in [M]$ with local parameter $\theta^{(i)}_{0,0} = \bar{\theta}_{0}$ ;
% \For{global iteration $r = 0,1,\ldots,R-1$}
%     \State $\rhd$ \textbf{Client side:}
%     \For{each client $i \in [M]$ in parallel}
%         \State $\theta^{(i)}_{r+1} = \texttt{NonlinearClientUpdate}\big(i, \bar{\theta}_{r}, K_i\big)$ (\ref{eq:fedavg})
        
%         \State Send $\theta^{(i)}_{r+1}$ to server
%     \EndFor
%     \State $\rhd$ \textbf{Server side:}
%     \State $\bar{\theta}_{r+1} = \frac{1}{M} \sum_{i=1}^M \theta^{(i)}_{r+1}$
    
%     \State Broadcast $\bar{\theta}_{r+1}$ to all clients
% \EndFor
% \State \textbf{Return} global estimate $\bar{\theta}_{R}$
% \end{algorithmic}
% \end{algorithm}
This approach reduces communication overhead by allowing multiple updates locally. 
After completing these updates, the clients send their updated local models $\bar{\theta}_{r+1}^{(i)}$ to the server. The server then averages all updates to obtain a new global model $\bar{\theta}_{r+1}$, which is subsequently broadcast back to all clients to synchronize (line 11). This  efficiently balances local computation and communication, addressing potentially limited communication capabilities between clients.
\begin{algorithm}
\caption{\texttt{FNSysId}: Federated Nonlinear System Identification}
\small
\label{FNSysId_algo}
\begin{algorithmic}[1]
\State \textbf{Initialize} server with global parameter $\bar{\theta}_{0}$ and learning rate $\alpha$ ;
\State \textbf{Initialize} each client $i \in [M]$ with local parameter $\theta^{(i)}_{0,0} = \bar{\theta}_{0}$ ;
\For{global iteration $r = 0,1,\ldots,R-1$}
    \State $\rhd$ \textbf{Client side:}
    \For{each client $i \in [M]$ in parallel}
        \State $\theta^{(i)}_{r+1} = \texttt{NonlinearClientUpdate}\big(i, \bar{\theta}_{r}, K_i\big)$ (\ref{eq:fedavg})
        
        \State Send $\theta^{(i)}_{r+1}$ to server
    \EndFor
    \State $\rhd$ \textbf{Server side:}
    \State $\bar{\theta}_{r+1} = \frac{1}{M} \sum_{i=1}^M \theta^{(i)}_{r+1}$
    
    \State Broadcast $\bar{\theta}_{r+1}$ to all clients
\EndFor
\State \textbf{Return} global estimate $\bar{\theta}_{R}$
\end{algorithmic}
\end{algorithm}
\begin{corollary}
    Using the same setup as Theorem \ref{theorem:1} and for all communication rounds $R\geq 1$, the output $\bar{\theta}_{R}$ given by the FNSysId algorithm with the FedAvg update rule satisfies: $$\max_{i \in [M]} \|\bar{\theta}_{R} - \theta^{(i)*}\|_2 \le \mathcal{O}\left(\frac{1}{KR} + \frac{1}{\sqrt{\sum_{i=1}^{M} N_{i} T}} + \epsilon\right).$$
\end{corollary}
\begin{proof}
    Because the system studied in this paper is linearly-parameterized, the proof directly follows from \cite[Corollary 4]{fedsysid}.
\end{proof}

\begin{figure*}[h]
\centering
\subfigure{
    \begin{minipage}[t]{0.28\textwidth}
    \centering
    \includegraphics[width=\textwidth,trim=0 0 0 30, clip]{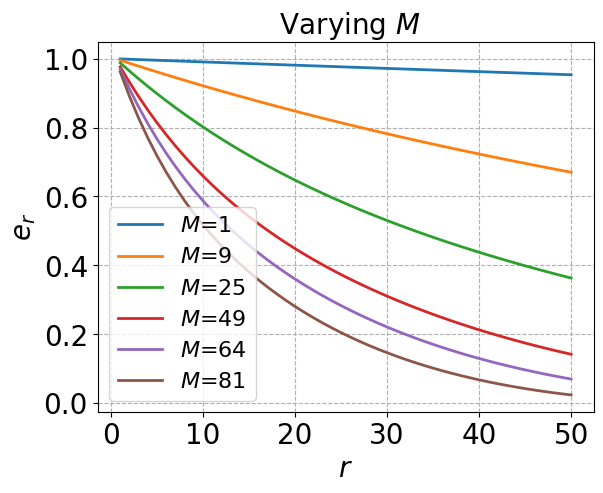}
    \vspace{1mm}
    % \textbf{[a]}
    (a)
    \label{fig:Pend_1_M}
    \end{minipage}
}
\hfill
\subfigure{
    \begin{minipage}[t]{0.28\textwidth}
    \centering
    \includegraphics[width=\textwidth,trim=0 0 0 30, clip]{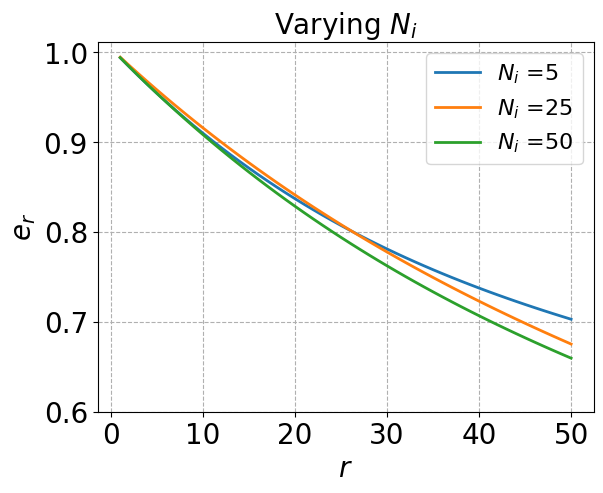}
    \vspace{1mm}
    % \textbf{[b]}
    (b)
    \label{fig:Pend_1_N}
    \end{minipage}
}
\hfill
\subfigure{
    \begin{minipage}[t]{0.28\textwidth}
    \centering
    \includegraphics[width=\textwidth,trim=0 0 0 30, clip]{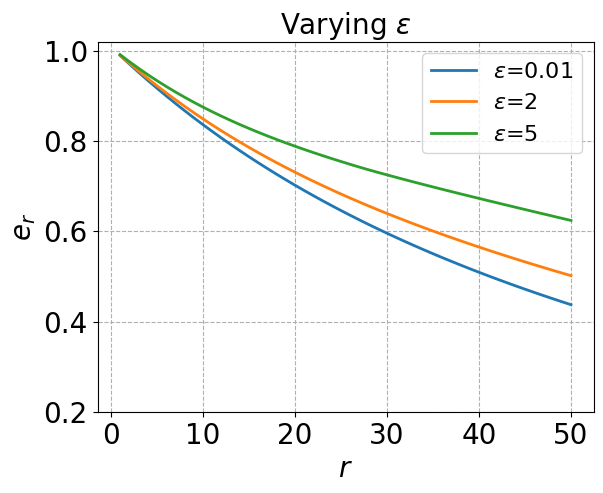}
    \vspace{1mm}
    % \textbf{[c]}
    (c)
    \label{fig:Pend_1_eps}
    \end{minipage}
}
\caption{Estimation error versus the number of global iterations for the real-world nonlinear dynamical system of a pendulum using gradient descent (GD). Results illustrate the impact of varying the number of clients ($M$), the number of local samples per client ($N_i$), and the heterogeneity parameter ($\epsilon$), with each client performing $K_i = 1$ local updates at alearning rate of $10^{-2}$. Subfigures: (a) $N_i = 10$, $\epsilon = 0.01$; (b) $M = 10$, $\epsilon = 0.01$; (c) $M = 20$, $N_i = 10$.}
\label{fig:pend_sy_id}
\end{figure*}

% \begin{figure*}[t]
% \centering
% \subfigure{
%     \begin{minipage}[t]{0.28\textwidth}
%     \centering
%     \includegraphics[width=\textwidth,trim=0 0 0 30, clip]{Pendulum_sys_id_SGD_1_M.png}
%     % \vspace{1mm}
%     (a)
%     \label{fig:Pend_SGD_M}
%     \end{minipage}
% }
% \hfill
% \subfigure{
%     \begin{minipage}[t]{0.28\textwidth}
%     \centering
%     \includegraphics[width=\textwidth,trim=0 0 0 0, clip]{Pendulum_sys_id_SGD_1_N.png}
%     % \vspace{1mm}
%     % \textbf{[b]}
%     (b)
%     \label{fig:Pend_SGD_N}
%     \end{minipage}
% }
% \hfill
% \subfigure{
%     \begin{minipage}[t]{0.28\textwidth}
%     \centering
%     \includegraphics[width=\textwidth,trim=0 0 0 0, clip]{Pendulum_sys_id_SGD_1_eps.png}
%     % \vspace{1mm}
%     % \textbf{[c]}
%     (c)
%     \label{fig:Pend_SGD_eps}
%     \end{minipage}
% }
% \caption{Estimation error vs. global iterations for the nonlinear pendulum system using mini-batch SGD (batch size $10$). Results illustrate the impact of varying the number of clients ($M$), the number of local samples per client ($N_i$), and the heterogeneity parameter ($\epsilon$), with each client performing $K_i = 2$ local updates at learning rate of $10^{-2}$. Subfigures: (a) $N_i = 10$, $\epsilon = 0.01$; (b) $M = 10$, $\epsilon = 0.01$; (c) $M = 20$, $N_i = 10$.}
% \label{fig:pend_sy_id_SGD}
% % \vspace{-0.5em}
% \end{figure*}

\begin{figure*}[t]
\centering
\subfigure{
    \begin{minipage}[t]{0.28\textwidth}
    \centering
    \includegraphics[width=\textwidth,trim=0 0 0 30, clip]{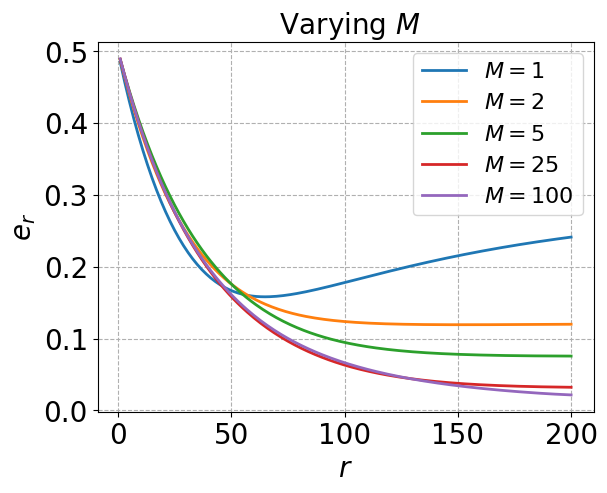}
    \vspace{0.5mm}
    % \textbf{[a]}
    (a)
    \label{fig:syn_M}
    \end{minipage}
}
\hfill
\subfigure{
    \begin{minipage}[t]{0.28\textwidth}
    \centering
    \includegraphics[width=\textwidth,trim=0 0 0 30, clip]{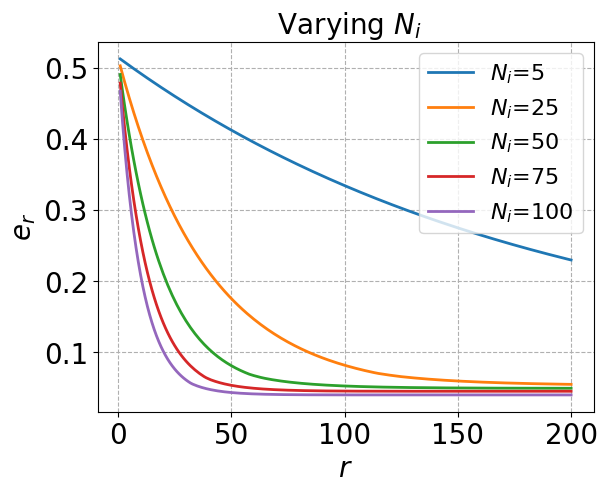}
    \vspace{0.5mm}
    % \textbf{[b]}
    (b)
    \label{fig:syn_N}
    \end{minipage}
}
\hfill
\subfigure{
    \begin{minipage}[t]{0.28\textwidth}
    \centering
    \includegraphics[width=\textwidth,trim=0 0 0 30, clip]{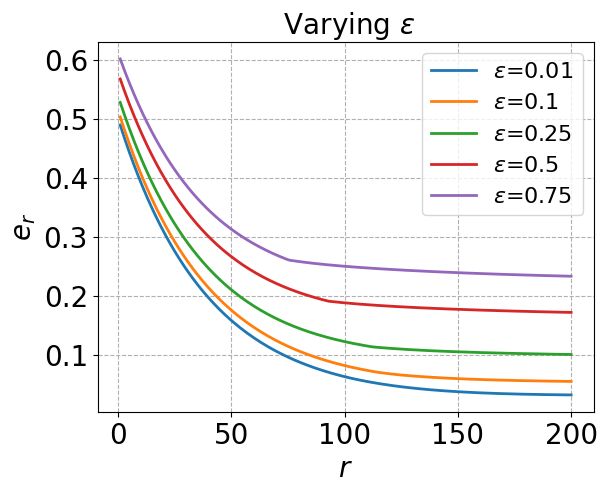}
    \vspace{0.5mm}
    % \textbf{[c]}
    (c)
    \label{fig:syn_eps}
    \end{minipage}
}
\caption{Estimation error on synthetic data as a function of global iterations using gradient descent, evaluated across different client configurations of the number of clients ($M$), local dataset size per client ($N_i$), and heterogeneity parameter ($\epsilon$). In all cases, each client performs $K_i = 5$ local update steps with a fixed learning rate of $10^{-4}$. The following configurations are considered:
(a) $N_i = 10$, $\epsilon = 0.1$ with varying $M$;
(b) $M = 25$, $\epsilon = 0.1$ with varying $N_i$;
(c) $M = 25$, $N_i = 25$ with varying $\epsilon$.}
\label{fig:syn_GD_K_5}
% \vspace{-0.5em}
\end{figure*}

% \begin{figure*}[t]
% \centering
% \subfigure{
%     \begin{minipage}[t]{0.28\textwidth}
%     \centering
%     \includegraphics[width=\textwidth,trim=0 0 0 30, clip]{pend_error_vs_clients_count.png}
%     \vspace{0.5mm}
%     % \textbf{[a]}
%     (a)
%     \label{fig:pend_error_vs_clients_count}
%     \end{minipage}
% }
% \hfill
% \subfigure{
%     \begin{minipage}[t]{0.28\textwidth}
%     \centering
%     \includegraphics[width=\textwidth,trim=0 0 0 30, clip]{synthetic_data_error_vs_clients_count.png}
%     \vspace{0.5mm}
%     % \textbf{[b]}
%     (b)
%     \label{fig:syn_error_vs_clients_count}
%     \end{minipage}
% }
% \hfill
% \subfigure{
%     \begin{minipage}[t]{0.28\textwidth}
%     \centering
%     \includegraphics[width=\textwidth,trim=0 0 0 30, clip]{e_vs_K.png}
%     \vspace{0.5mm}
%     % \textbf{[b]}
%     (c)
%     \label{fig:e_vs_local_updates}
%     \end{minipage}
% }
% \caption{Comparison of estimation error versus $\sqrt{M}$  on (a) a real-world pendulum system and (b) synthetic data. The empirical results validate that, in low heterogeneity settings, the non-asymptotic convergence rate can be enhanced by increasing number of clients. (c) Impact of local updates $(K_i)$ on  estimation error.}
% \label{fig:error_vs_clients_count}
% \end{figure*}
\section{Experiments}\label{sec:experiments}

\subsection{Data}
In this section we describe the datasets used in our experiments.
\subsubsection{Synthetic Data} 
To evaluate the performance and efficiency of \texttt{FNSysId}, we conduct experiments involving $M$ clients, each associated with a discrete-time dynamical system characterized by $n = 3$ states and $p = 2$ inputs. The initial state, input signal, and process noise are all sampled from zero-mean distributions with unit standard deviations. The trajectories were generated by augmenting the trajectory formulation presented by  \cite{fedsysid} with a nonlinear term, as presented in the following equation.
\begin{align}
    x_{t+1}^{(i)} = A^{(i)} \sin(x_{t}^{(i)}) + B^{(i)} u_{t}^{(i)} + w_{t}^{(i)}.
\end{align}
Following the approach of \cite{xin2022identifying}, we construct heterogeneous client dynamics by perturbing a nominal system $(A_0, B_0)$. Particularly, for client $i$, the perturbed system matrices are as follows:
\begin{equation}
    A^{(i)} = A_0 + \gamma_1^{(i)} V,
    \label{eq:modify_A}
    \end{equation}
    \begin{equation}
    B^{(i)} = B_0 + \gamma_2^{(i)} U,
    \label{eq:modify_B}
\end{equation}
where $\gamma_1^{(i)}\sim \mathrm{U}(0, \epsilon)$ and $\gamma_2^{(i)} \sim \mathrm{U}(0, \epsilon)$ are uniformly distributed random variables.
% \red{Max: You're saying the $\gamma_j^{(i)}\sim U(0,e)$ but then that $\gamma_j=\epsilon r^{(i)}$. Is $r^{(i)}\in[0,1]$. If so, we can remove the second definition since that is clear from the distribution.}
% Here $\gamma_1=\epsilon \cdot r^{(i)} , \quad i = 1, 2, \ldots, M  $ and
% $\gamma_2=\epsilon \cdot r^{(i)} , \quad i = 1, 2, \ldots, M$ where each $r^{(i)}$ represents some independent random number.
% where $V \in \mathbb{R}^{3 \times 3}$ and $U \in \mathbb{R}^{3 \times 2}$ define fixed perturbation directions. The specific matrix values are:
% % A_0 = np.array([[1, 0.2, 0.6],
% %                     [0.1, 0.4, 0.4],
% %                     [0.2, 0.3, 0.4]], dtype=np.float64, order='F')
% % B_0 = np.array([[0.6, 0.8],
% %                     [1.0, 1.0],
% %                     [0.6, 0.5]], dtype=np.float64, order='F')
% % V = np.diag([0, 1, 1]).astype(np.float64, order='F')
% % U = np.array([[1, 0], [1, 0], [0, 1]], dtype=np.float64, order='F')

% \[
% A_0 = \begin{bmatrix}
% 1 & 0.2 & 0.6 \\
% 0.1 & 0.4 & 0.4 \\
% 0.2 & 0.3 & .4
% \end{bmatrix}, \quad
% V = \begin{bmatrix}
% 0.6 & 0.5 & 0.4 \\
% 0 & 1 & 0 \\
% 0 & 0 & 1
% \end{bmatrix}
% \]
% \[
% B_0 = \begin{bmatrix}
% 0.6 & 0.8 \\
% 1 & 1 \\
% 0.6 & 0.5
% \end{bmatrix}, \quad
% U = \begin{bmatrix}
% 1 & 0 \\
% 1 & 0 \\
% 0 & 1
% \end{bmatrix}
% \]

We consider a trajectory length of $T = 5$ and assume that each client uses an equal number of data points.
% After computing the aggregated model $\bar{\Theta}$ using the FedSysID algorithm (as defined in Eq.~(4)), we quantify the estimation performance by computing the average relative $\ell_2$-norm error between the estimated and true parameters at each global round $r$.

\subsubsection{Real-World System} 
We conduct experiments across two different physical systems. The
pendulum system is characterized by one state and one input, targeting the estimation of two unknown parameters\footnote{Details on the trajectory generation method are in \cite [Section 4]{musavi2024identification}}. In all both experiments, the control input is defined as $u_t = \pi(x_t) + \eta_t,$ where \( \pi(x_t) \) denotes the control policy evaluated at state \( x_t \) and \( \eta_t \) represents zero-mean i.i.d.\ noise, sampled. Here \( \eta_t \) is sampled from  a uniform distribution.

We have two nominal parameters:
$A_0 = \tfrac{1}{l}$ and $B_0 = \tfrac{1}{ml^{2}}$ (\( m \) denotes the mass of the pendulum and \( l \) represents its length). Following the formulation of \cite{musavi2024identification}, we modified the trajectories by introducing the heterogeneity using the same procedure outlined in equations \eqref{eq:modify_A} and \eqref{eq:modify_B}.
\begin{equation}
    \ddot{\alpha} = -A^{(i)} g \sin(\alpha) + B^{(i)}u + w ,
\end{equation}
After discretization, the system dynamics can be rewritten in the structure of (\ref{eq:system}).

In the quadrotor example presented in~\cite[Section 4]{musavi2024identification}, the control input is specified as $u_t = \pi(x_t) + \eta_t$, where $\pi(x_t)$ denotes the nominal controller designed according to the approach in~\cite{alaimo2013mathematical}. The system comprises $13$ states and $4$ control inputs. The unknown parameter matrix $\theta^{*}$ includes $7$ parameters to learn.

\subsection{Estimation Error}
For each client $C_i$, the normalized error between the client and the server parameters is calculated as:
\begin{equation}
e_r^{(i)} = \frac{\left\|\theta_{s,r} -  \theta^{(i)*}  \right\|_2}{\left\| \theta^{(i)*} \right\|_2}.
\label{eq:error_per_client}
\end{equation}

The maximum estimation error reported for communication round $r$ is:
$e_{r} = \max_{1 \leq i \leq M} e_r^{(i)}$, which also serves as a heterogeneity metric that quantifies how far the global model is from the most divergent client.
\begin{figure*}[t]
\centering
\subfigure{
    \begin{minipage}[t]{0.28\textwidth}
    \centering
    \includegraphics[width=\textwidth,trim=0 0 0 30, clip]{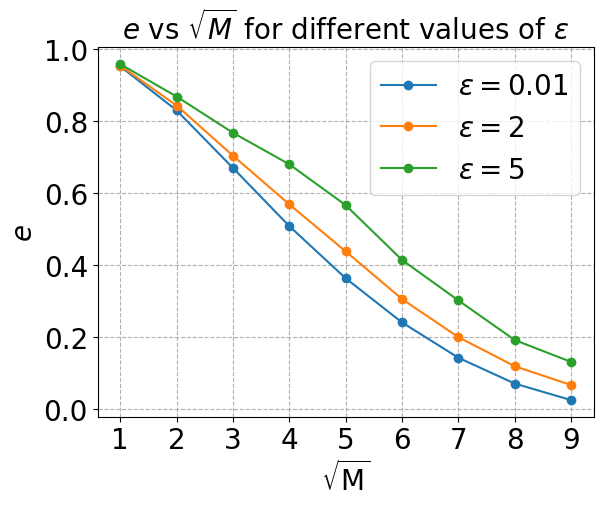}
    \vspace{0.5mm}
    % \textbf{[a]}
    (a)
    \label{fig:pend_error_vs_clients_count}
    \end{minipage}
}
\hfill
\subfigure{
    \begin{minipage}[t]{0.28\textwidth}
    \centering
    \includegraphics[width=\textwidth,trim=0 0 0 30, clip]{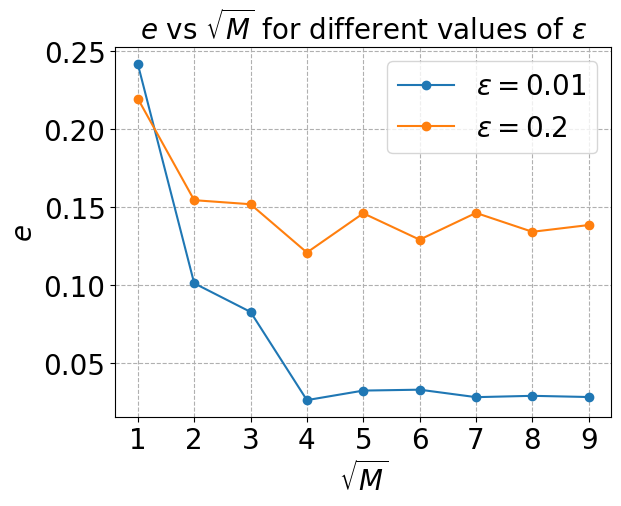}
    \vspace{0.5mm}
    % \textbf{[b]}
    (b)
    \label{fig:syn_error_vs_clients_count}
    \end{minipage}
}
\hfill
\subfigure{
    \begin{minipage}[t]{0.28\textwidth}
    \centering
    \includegraphics[width=\textwidth,trim=0 0 0 30, clip]{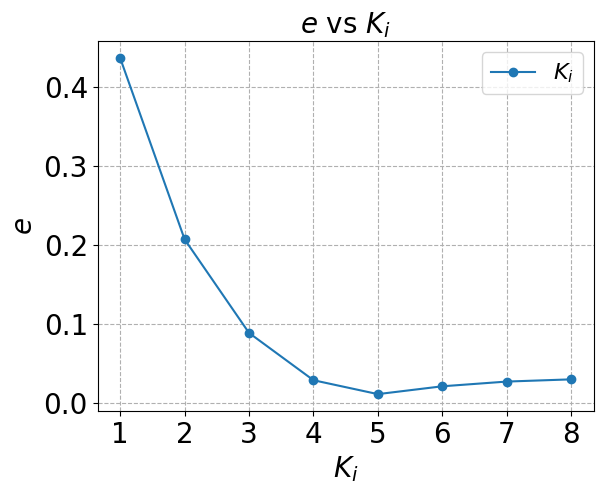}
    \vspace{0.5mm}
    % \textbf{[b]}
    (c)
    \label{fig:e_vs_local_updates}
    \end{minipage}
}
\caption{Comparison of estimation error versus $\sqrt{M}$  on (a) a real-world pendulum system and (b) synthetic data. The empirical results validate that, in low heterogeneity settings, the non-asymptotic convergence rate can be enhanced by increasing number of clients. (c) Impact of local updates $(K_i)$ on  estimation error.}
\label{fig:error_vs_clients_count}
\end{figure*}
\section{Results}
Figure  \ref{fig:pend_sy_id} demonstrates the experimental results the pendulum.
In Figure \ref{fig:Pend_1_M}, as the number of clients increases, the estimation error decays faster.
Figure \ref{fig:Pend_1_N} shows that data volumme per client per round slighly improves the convergence rate. Figure \ref{fig:Pend_1_eps} illustrates the adverse impact of system heterogeneity on convergence behavior. Under a fixed number of clients ($M_i=10$) and trajectories ($N_i=10$), increasing the heterogeneity parameter the estimation error. Additional results for the quadrotor system can be found in the Appendix.

% Figure~\ref{fig:pend_sy_id_SGD} demonstrates the system behavior under stochastic gradient descent (SGD). Compared to ~\ref{fig:pend_sy_id}, SGD improves the convergence rate. Additional results for the quadrotor system have been provided in the Appendix.

Figure~\ref{fig:syn_GD_K_5} presents empirical convergence results for on synthetic datasets. Figure~\ref{fig:syn_M} illustrates that increasing the number of clients enhances estimation accuracy, Figure~\ref{fig:syn_N} demonstrates that increasing the number of trajectories per client improves estimation quality. Conversely, Figure~\ref{fig:syn_eps} indicates a degradation in estimation performance as the heterogeneity parameter increases.

Figure~\ref{fig:error_vs_clients_count} shows how the normalized estimation error ($e$) varies with the square root of the number of clients ($\sqrt{M}$) for the pendulum dynamics and synthetic data. This result validates that the non-asymptotic convergence rate proposed is empirically consistent in the low heterogeneity regime.

Figure~\ref{fig:e_vs_local_updates} illustrates that increasing the number of local updates per communication round can reduce communication overhead by lowering the frequency of global aggregations. However, beyond a certain point, the performance gains diminish as excessive local training in each global iteration introduces significant global model divergence across clients. 
% The experimental setup is similar to that of Figure~\ref{fig:pend_sy_id_SGD}, except for a variation in the number of local updates.

\section{Conclusion}
\iffalse We introduced a federated learning framework for linearly-parameterized nonlinear system identification. We derived tight finite-sample error bounds and validated our theory on the pendulum and quadrotor nonlinear physical systems. We experimentally showed that the least squares estimator enjoys a $\sqrt{M}$ convergence rate improvement over the centralized case. In summary, we provided the first theoretical and experimental study on federated nonlinear system identification.\fi

We introduced a federated learning framework for identifying linearly-parameterized nonlinear dynamical systems, with a particular focus on piecewise affine (PWA) models. Our theoretical analysis shows that the convergence error decreases as $1/\sqrt{M}$, where $M$ is the number of clients, thus having significant improvement in convergence as more clients collaborate. We corroborate our theory on the pendulum and quadrotor nonlinear physical systems, experimentally demonstrating the improvement in convergence error at any client decreases favorably as more clients collaborate for federated system identification. 

Our work opens many interesting future directions. Theoretical analysis of how different optimization hyperparameters (e.g., number of local epochs in each global iteration) impact convergence is of significant interest. Adaptive batch size strategies could also balance the trade-off between gradient noise and convergence speed. A key fundamental open problem is when $\phi$ is not known a priori. \iffalse In such cases, it is critical to evaluate how different choices of $\phi$ affect convergence. \fi One could explore scenarios where $\phi$ itself is learned collaboratively alongside the model, forming an end-to-end variant of our proposed framework. Finally, while this work focused on PWA models, a natural extension is to consider linearly-parameterized nonlinear systems via Koopman theory \cite{koopmantheory}, which is based on lifting the states into an infinite-dimensional feature space, where the dynamics evolve linearly. While exact infinite-dimensional embeddings are impractical, one can explore learning a finite set of invariant eigenfunctions to approximate this transformation. A formal proof establishing that Algorithm 1 converges to the quantity characterized in Theorem 1 remains an important direction for future work. While empirical results demonstrate consistent convergence behavior, deriving rigorous theoretical guarantees is ongoing. In particular, future work will focus on establishing convergence guarantees for both GD and SGD settings, including sufficient conditions and convergence rates.

\section*{Acknowledgment}
This work is supported in part by ANRF under PM ECRG Grant (ANRF/ECRG/2024/005912/ENS), in part by Wadhwani School of Data Science \& AI (WSAI) Exploratory Research Grant (SB25261132EEWDWI009167), in part by Centre for Responsible AI (CeRAI), and in part by IIT Madras under New Faculty Initiation Grant (RF24251118EENFIG009167). The authors thank Nandan Sudarsanam for fruitful discussions.

\begingroup
\setlength{\itemsep}{0pt}
\setlength{\parskip}{0pt}
\setlength{\parsep}{0pt}
\bibliographystyle{IEEEtran}
\bibliography{IEEEexample}

@article{mania2022active,
  title={Active learning for nonlinear system identification with guarantees},
  author={Mania, Horia and Jordan, Michael I. and Recht, Benjamin},
  journal={J. Machine Learning Research},
  volume={23},
  number={32},
  pages={1--30},
  year={2022}
}

@incollection{willems1989models,
  title={Models for dynamics},
  author={Willems, Jan C.},
  booktitle={Dynamics Reported},
  pages={171--269},
  year={1989},
  key={Springer},
}

@book{guckenheimer2013nonlinear,
  title={Nonlinear Oscillations, Dynamical Systems, and Bifurcations of Vector Fields},
  author={Guckenheimer, John and Holmes, Philip},
  year={2013},
  publisher={Springer}
}

@article{sarkar2021finite,
  title={Finite time {LTI} system identification},
  author={Sarkar, Tuhin and Rakhlin, Alexander and Dahleh, Munther A.},
  journal={J. Machine Learning Research},
  volume={22},
  number={26},
  pages={1--61},
  year={2021}
}

@article{venkatesh2002system,
  title={On system identification of complex systems from finite data},
  author={Venkatesh, Saligrama R. and Dahleh, Munther A.},
  journal={IEEE Trans. Automatic Control},
  volume={46},
  number={2},
  pages={235--257},
  year={2002}
}

@incollection{
musavi2024identification,
title={Identification of analytic nonlinear dynamical systems with non-asymptotic guarantees},
author={Negin Musavi and Ziyao Guo and Geir Dullerud and Yingying Li},
booktitle={Advances in Neural Information Processing Systems},
year={2024},
volume = "37",
 pages = {85500--85522}
}

@inproceedings{fedsysid,
    author = {Han Wang and Leonardo F. Toso and James Anderson},
    title = {{FedSysID}: A federated approach to sample-efficient system identification},
    booktitle = {Proc. 5th Annu. Learning Dynamics Control Conf.},
    year = "2023",
    pages = "1308--1320"
}

@InProceedings{pmlr-v75-simchowitz18a,
  title = 	 {Learning without mixing: Towards a sharp analysis of linear system identification},
  author =       {Simchowitz, Max and Mania, Horia and Tu, Stephen and Jordan, Michael I. and Recht, Benjamin},
  booktitle = 	 {Proc. 31st Conf. Learning Theory},
  pages = 	 {439--473},
  year = 	 {2018}
}

@book{ljung1998system,
  title={System Identification},
  author={Ljung, Lennart},
  year={1998},
  publisher={Pearson}
}

@inproceedings{JedP20,
  title={Finite-time identification of stable linear systems optimality of the least-squares estimator},
  author={Jedra, Yassir and Proutiere, Alexandre},
  booktitle={Proc. 59th IEEE Conf. Decision Control (CDC)},
  pages={996--1001},
  year={2020}
}

@inproceedings{sarkar2019near,
  title={Near optimal finite time identification of arbitrary linear dynamical systems},
  author={Sarkar, Tuhin and Rakhlin, Alexander},
  booktitle={Proc. 36th Int. Conf. Machine Learning},
  pages={5610--5618},
  year={2019}
}

@inproceedings{mcmahan2023communicationefficientlearningdeepnetworks,
      title={Communication-efficient learning of deep networks from decentralized data}, 
      author={H. Brendan McMahan and Eider Moore and Daniel Ramage and Seth Hampson and Agüera y Arcas, Blaise},
    booktitle = "Proc. 20th Int. Conf. Artificial Intelligence Statistics", 
    pages = "1273--1282",
    year = "2017"
}

@article{Jiang_2019,
   title={Irrelevance of linear controllability to nonlinear dynamical networks},
   volume={10},
   DOI={10.1038/s41467-019-11822-5},
   number={1},
   journal={Nature Communications},
   publisher={Springer Science and Business Media LLC},
   author={Jiang, Junjie and Lai, Ying-Cheng},
   year={2019},
   month=sep }

@inproceedings{SibaiH2018,
author = {Sibai, Hussein and Mitra, Sayan},
title = {State estimation of dynamical systems with unknown inputs: Entropy and bit rates},
year = {2018},
booktitle = {Proc. 21st Int. Conf. Hybrid Syst.: Computation and Control},
pages = {217--226}
}

@inproceedings{truex19,
  title={A hybrid approach to privacy-preserving federated learning},
  author={Truex, Stacey and Baracaldo, Nathalie and Anwar, Ali and Steinke, Thomas and Ludwig, Heiko and Zhang, Rui and Zhou, Yi},
  booktitle={Proc. 12th ACM Workshop Artificial Intelligence and Security},
  pages={1--11},
  year={2019}
}

@misc{kon16,
  title={{Federated learning: Strategies for improving communication efficiency}},
  author={Kone{\v{c}}n{\`y}, Jakub and McMahan, H Brendan and Yu, Felix X. and Richt{\'a}rik, Peter and Suresh, Ananda Theertha and Bacon, Dave},
  howpublished={arXiv:1610.05492},
  year={2016}
}

@inproceedings{Sahu2018OnTC,
  title={Federated optimization in heterogeneous networks},
  author={Li, Tian and Sahu, Anit Kumar and Zaheer, Manzil and Sanjabi, Maziar and Talwalkar, Ameet and Smith, Virginia},
 booktitle = {Proc. Machine Learning Syst.},
  year={2020},
 pages = {429--450}
}

@article{yang2019federated,
  title={Federated learning},
  author={Yang, Qiang and Liu, Yang and Cheng, Yong and Kang, Yan and Chen, Tianjian and Yu, Han},
  journal={Synthesis Lectures on Artificial Intelligence and Machine Learning},
  volume={13},
  number={3},
  pages={1--207},
  year={2019},
  publisher={Morgan \& Claypool Publishers}
}

@ARTICLE{11077727,
  author={Chen, Yutao and Chen, Wei},
  journal={IEEE Trans. Signal Processing}, 
  title={Kalman Filter Aided Federated {K}oopman Learning}, 
  year={2025},
  volume={73},
  number={},
  pages={2879-2895},
  doi={10.1109/TSP.2025.3587329}}

@article{koopmantheory,
author = {Brunton, Steven L. and Budi\v{s}i\'{c}, Marko and Kaiser, Eurika and Kutz, J. Nathan},
title = "Modern {K}oopman theory for dynamical systems",
journal = {SIAM Rev.},
volume = {64},
number = {2},
pages = {229--340},
year = {2022},
doi = {10.1137/21M1401243},
}

@inproceedings{xin2022identifying,
  title={Identifying the dynamics of a system by leveraging data from similar systems},
  author={Xin, Lei and Ye, Lintao and Chiu, George and Sundaram, Shreyas},
  booktitle={Proc. American Control Conf. (ACC)},
  pages={818--824},
  year={2022},
}

@inproceedings{alaimo2013mathematical,
  title={Mathematical modeling and control of a hexacopter},
  author={Alaimo, A. and Artale, V. and Milazzo, C. and Ricciardello, A. and Trefiletti, L.},
  booktitle={Proc. 2013 Int. Conf. Unmanned Aircraft Syst. (ICUAS)},
  pages={1043--1050},
  year={2013},
}

@article{Kellett2014Compendium,
  title={A compendium of comparison function results},
  author={Kellett, Christopher M.},
  journal={Mathematics of Control, Signals, and Systems},
  volume={26},
  number={3},
  pages={339--374},
  year={2014},
  publisher={Springer}
}

@BOOK{NAP26894,
  author    = {{National Academy of Engineering and National Academies of Sciences, Engineering, and Medicine}},
  title     = "Foundational Research Gaps and Future Directions for Digital Twins",
  isbn      = "978-0-309-70042-9",
  doi       = "10.17226/26894",
  url       = "https://nap.nationalacademies.org/catalog/26894/foundational-research-gaps-and-future-directions-for-digital-twins",
  year      = 2024,
  publisher = "The National Academies Press",
  address   = "Washington, DC"
}
\endgroup

%%%%%%%%%%%%%%%%%%%%%%%%%%%%%%%%%%%%%%%%%%%%%%%%%%%%%%%%%%%%%%%%%%%%%%%%%%%%%%%%
% \section*{Appendix}
% \newpage
\appendix
\begin{figure*}[t]
\centering
\subfigure{
    \begin{minipage}[t]{0.28\textwidth}
    \centering
    \includegraphics[width=\textwidth,trim=0 0 0 30, clip]{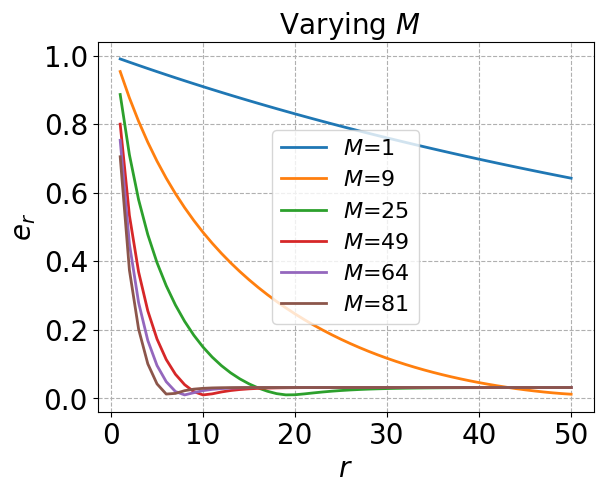}
    % \vspace{1mm}
    (a)
    \label{fig:Pend_SGD_M}
    \end{minipage}
}
\hfill
\subfigure{
    \begin{minipage}[t]{0.28\textwidth}
    \centering
    \includegraphics[width=\textwidth,trim=0 0 0 0, clip]{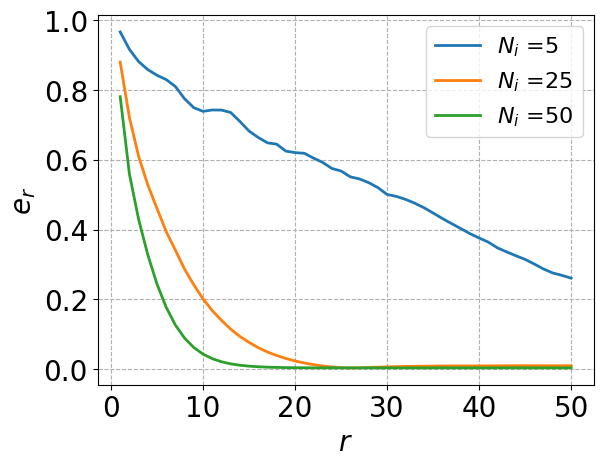}
    % \vspace{1mm}
    % \textbf{[b]}
    (b)
    \label{fig:Pend_SGD_N}
    \end{minipage}
}
\hfill
\subfigure{
    \begin{minipage}[t]{0.28\textwidth}
    \centering
    \includegraphics[width=\textwidth,trim=0 0 0 0, clip]{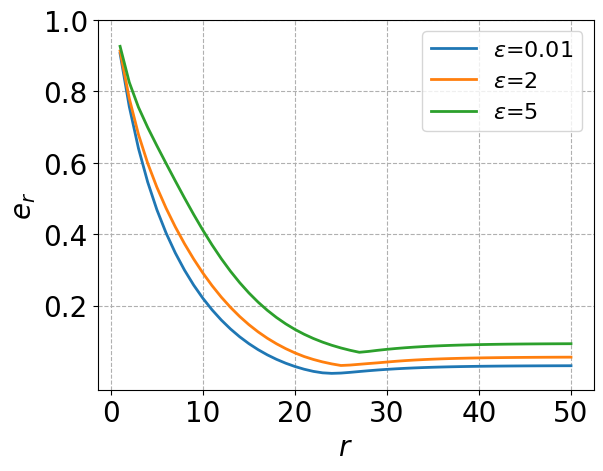}
    % \vspace{1mm}
    % \textbf{[c]}
    (c)
    \label{fig:Pend_SGD_eps}
    \end{minipage}
}
\caption{Estimation error vs. global iterations for the nonlinear pendulum system using mini-batch SGD (batch size $10$). Results illustrate the impact of varying the number of clients ($M$), the number of local samples per client ($N_i$), and the heterogeneity parameter ($\epsilon$), with each client performing $K_i = 2$ local updates at learning rate of $10^{-2}$. Subfigures: (a) $N_i = 10$, $\epsilon = 0.01$; (b) $M = 10$, $\epsilon = 0.01$; (c) $M = 20$, $N_i = 10$.}
\label{fig:pend_sy_id_SGD}
% \vspace{-0.5em}
\end{figure*}

\begin{figure*}[!t]
\centering
\subfigure{
    \begin{minipage}[t]{0.28\textwidth}
    \centering
    \includegraphics[width=\textwidth,trim=0 0 0 30, clip]{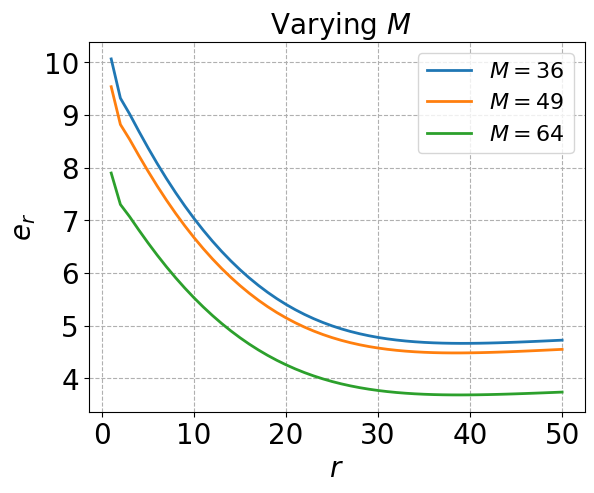}
    \vspace{1mm}
    % \textbf{[a]}
    (a)
    \label{fig:quad_1_M}
    \end{minipage}
}
\hspace{0.05\textwidth}
\subfigure{
    \begin{minipage}[t]{0.28\textwidth}
    \centering
    \includegraphics[width=\textwidth,trim=0 0 0 30, clip]{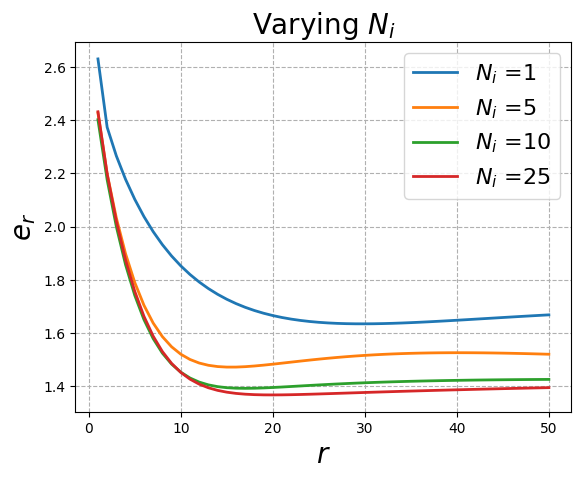}
    \vspace{1mm}
    % \textbf{[b]}
    (b)
    \label{fig:quad_1_N}
    \end{minipage}
}
\caption{Estimation error versus the number of global iterations for the real-world nonlinear dynamical system of a quadrotor using gradient descent (GD). Results illustrate the impact of varying the number of clients ($M$), the number of local samples per client ($N_i$),with each client performing $K_i = 5$ local updates at a learning rate of $10^{-1}$. Subfigures: (a) $N_i = 10$, $\epsilon = 2$; (b) $M = 5$, $\epsilon =5$.}
\label{fig:quad_sy_id}
\end{figure*}
\label{sec:appendix}
In this section, we provide definitions for the $\mathcal{K}$ and $\mathcal{KL}$ functions, proofs for Lemma \ref{lemma:1}, Proposition \ref{prop:1}, Proposition \ref{prop:2}, and Theorem \ref{theorem:1}. In the end, we provide the results for quadrotor experiments.
\begin{definition}[Classes $\mathcal{K}$, and $\mathcal{KL}$]
\label{defclasses}
The following function class definitions are cited from \cite{Kellett2014Compendium}.
\begin{enumerate}
\item A function $\alpha:[0,a)\to[0,\infty)$ (with $a\in(0,\infty]$) is of class $\mathcal{K}$ if it is continuous, strictly increasing, and $\alpha(0)=0$.

\item A function $\beta:[0,a)\times[0,\infty)\to[0,\infty)$ is of class $\mathcal{KL}$ if, for each fixed $t\ge 0$, the map $r\mapsto \beta(r,t)$ belongs to $\mathcal{K}$, and for each fixed $r\in[0,a)$, the map $t\mapsto \beta(r,t)$ is continuous, nonincreasing, and $\lim_{t\to\infty}\beta(r,t)=0$.
\end{enumerate}
\end{definition}

\begin{definition}
    (BMSB \cite{pmlr-v75-simchowitz18a}). Let $\{\mathcal{F}_t\}_{t \ge 1}$ denote a filtration and let $\{y_t\}_{t \ge 1}$ be an $\{\mathcal{F}_t\}_{t \ge 1}$-adapted random process taking values in $\mathbb{R}^{n_y}$. We say $\{y_t\}_{t \ge 1}$ satisfies the $(k, \Gamma_{sb}, p)$-block martingale small-ball (BMSB) condition for a positive integer $k$, a $\Gamma_{sb} \succ 0$, and a $p \in [0, 1]$, if for any fixed $v \in \mathbb{R}^{n_y}$ such that $\|v\|_2 = 1$, the process $\{y_t\}_{t \ge 1}$ satisfies $\frac{1}{k} \sum_{i=1}^k \mathbb{P}(|v^\top y_{t+i}| \ge \sqrt{v^\top \Gamma_{sb} v} \mid \mathcal{F}_t) \ge p$ almost surely for any $t \ge 1$.
    \label{def:bmsb}
\end{definition}

\subsection{Proof of Lemma 1}
\begin{proof}
    See~\cite[Theorem 1]{musavi2024identification}. This lemma simply changes notation to fit the federated setting.
\end{proof}

\subsection{Proof of Proposition 1}
\begin{proof}
    Fix a client $i$ and a unit vector $v\in\mathbb S^{n_\phi-1}$, and let $Y_t
:= v^\top\phi(x^{(i)}_t,u^{(i)}_t)$ for $t=0,\ldots,T_i-1$.  By Lemma 1
the sequence $\{Y_t\}$ satisfies the $(1,s_\phi,p_\phi)$
block–martingale small-ball condition, so Proposition 2.5 of
\cite{pmlr-v75-simchowitz18a}, with $k=1$ and $\alpha=1$, gives
\[
\Pr\!\bigl[\textstyle\sum_t Y_t^2 \le
           s_\phi^{2}\!\bigl(N_i p_\phi - 2\log(1/\delta)\bigr)\bigr]
     \,\le\,\delta.
\]
Take a $1/4$-net $\mathcal N$ of the unit sphere and apply the same bound to each
$v\in\mathcal N$ with failure level $\delta/|\mathcal N|:=\delta$; a union bound
gives that, with probability $\ge 1-\delta$, the inequality holds for every
$v\in\mathcal N$.  Because any unit $u$ is within $1/4$ of some
$v\in\mathcal N$
\[
\lambda_{\min}\bigl(\Phi^{(i)}\Phi^{(i)\!\top}\bigr) \;\ge\;
\frac12\,s_\phi^{2}\bigl(N_iT p_\phi - 2\log(9^{n_\phi}/\delta)\bigr).
\]
If $N_iT \ge \tfrac{4}{p_\phi}
       \bigl[n_\phi\log 9 + \log(1/\delta)\bigr]$, the component inside the bracket
is at least $\tfrac12 N_iT p_\phi$, and replacing $p_\phi$ by~$1$
loses only another factor $2$, giving
$\lambda_{\min}(\Phi^{(i)}\Phi^{(i)\!\top})
      \ge \tfrac12 s_\phi^{2}N_iT$
with probability at least $1-\delta$.
Finally, substituting $\delta/M$ for $\delta$ and union-bounding over the
$M$ clients yields the solution.
\end{proof}
\subsection{Proof of Proposition 2}
\begin{proof}
Conditioned on $\Phi^{(i)}$, the columns $w^{(i)}_t$
are independent, zero-mean, $\sigma_w$-sub-Gaussian vectors.
For any $t=0,\dots,T-1$, the matrix-form Bernstein inequality gives
\begin{align}
\label{eq:prop2}
\Pr\!\Bigl(
      \bigl\|W^{(i)}(\Phi^{(i)})^{T}\bigr\|_2 > t
    \Bigr)
    \;\le\;
    2\exp\!\Bigl(
       -\frac{t^{2}}{2\sigma_w^{2}\,\|\Phi^{(i)}\|_{F}^{2}}
    \Bigr).
\end{align}
Since $\|\phi(x,u)\|_2^{2}\le b_\phi$, we have
$\|\Phi^{(i)}\|_{F}^{2}\le N_iT\,b_\phi\le N_iT(n_x+n_\phi)$.
Choosing
\[
t \;=\;
4\sigma_w
\sqrt{\,N_iT\bigl(n_x+n_\phi+\log(2M/\delta)\bigr)}
\]
makes the right side of (\ref{eq:prop2}) less than or equal to $ \delta/(2M)$.  A union bound over
$i\in [M]$ and some algebra completes the proof.
\end{proof}

\subsection{Proof of Theorem 1}
\begin{proof}
Consider the relation for client $i$
\[
X^{(i)}_{+} \;=\; \theta^{(i)*}\,\Phi^{(i)} + W^{(i)} .
\]
Subtracting $\bar\theta_{\mathrm{LSE}}\Phi^{(i)}$ and regrouping gives
\[
X^{(i)}_{+}-\bar\theta_{\mathrm{LSE}}\Phi^{(i)}
   = (\theta^{(i)*}-\bar\theta_{\mathrm{LSE}})\,\Phi^{(i)} + W^{(i)} .
\]
After some algebra and defining \footnote{Because $\bar\theta_{\mathrm{LSE}}$ minimizes the pooled least‑squares
objective, the left side sum vanishes.} 
\(
G := \sum_{i}\Phi^{(i)}(\Phi^{(i)})^{\top}
\)
and
\(
\bar\theta^{*} := \tfrac1M\sum_{i}\theta^{(i)*},
\)
we obtain
\begin{align*}
G\,(\bar\theta_{\mathrm{LSE}}-\bar\theta^{*})^{\top} \!=\!
-\sum_{i=1}^M(\theta^{(i)*}-\bar\theta^{*}) \Phi^{(i)}(\Phi^{(i)})^{\top} \!- \!W^{(i)}(\Phi^{(i)})^{\top}.\nonumber
\end{align*}
We can bound each term as follows:
\begin{align*}
G\succeq\tfrac12\,s_\phi^{2}N_{\mathrm{tot}}\,I 
\end{align*}

by Proposition \ref{prop:1},

\begin{align*}
    \sum_{i}^MW^{(i)}(\Phi^{(i)})^T\leq4\sigma_w\sqrt{N_iT(n_x+n_\phi+log(2M/\delta))}
\end{align*}

by Proposition \ref{prop:2}, and
\begin{align*}
   ||\sum_i(\theta^{(i)*}-\bar{\theta}^*)\Phi^{(i)}(\Phi^{(i)})^\top||_2\leq \epsilon b_\phi N_iT
\end{align*}
using Assumption 5 \footnote{Notice that Assumption 5 implies that $||\theta^{(i)*}-\bar{\theta}^{(i)*}|| \leq \epsilon$, $\forall i\in [M]$.} and that $||\Phi^{(i)}(\Phi^{(i)})^\top||_2\leq N_ib_\phi$.

By multiplying both sides by $G^{-1}$, applying the bounds, taking the norm of each side gives with probability greater than or equal to $ 1-3\delta$,
\[
||\bar{\theta}_{LSE}-\bar{\theta}^* ||_2\leq \frac{8\sigma_w}{s_\phi^2}\sqrt{\frac{n_x+n_\phi+log(2M/\delta)}{\sum_{i=1}^MN_iT}}+\frac{b_\phi}{s_\phi^2}\epsilon.
\]
\medskip
\noindent
Since Assumption 5 implies
$\|\theta^{(i)*}-\bar\theta^{*}\|_2 \le \varepsilon/2$ for $i\in[M]$,
\[
\|\bar\theta_{\mathrm{LSE}}-\theta^{(i)*}\|_2
   \le \|\bar\theta_{\mathrm{LSE}}-\bar\theta^{*}\|_2 + \varepsilon/2 .
\]
This yields the expected bound.
\end{proof}
% \newpage
% \begin{figure*}[!t]
% \centering
% \subfigure{
%     \begin{minipage}[t]{0.28\textwidth}
%     \centering
%     \includegraphics[width=\textwidth,trim=0 0 0 30, clip]{quad_sys_id_GD_M.png}
%     \vspace{1mm}
%     % \textbf{[a]}
%     (a)
%     \label{fig:quad_1_M}
%     \end{minipage}
% }
% \hspace{0.05\textwidth}
% \subfigure{
%     \begin{minipage}[t]{0.28\textwidth}
%     \centering
%     \includegraphics[width=\textwidth,trim=0 0 0 30, clip]{quad_sys_id_GD_N.png}
%     \vspace{1mm}
%     % \textbf{[b]}
%     (b)
%     \label{fig:quad_1_N}
%     \end{minipage}
% }
% \caption{Estimation error versus the number of global iterations for the real-world nonlinear dynamical system of a quadrotor using gradient descent (GD). Results illustrate the impact of varying the number of clients ($M$), the number of local samples per client ($N_i$),with each client performing $K_i = 5$ local updates at a learning rate of $10^{-1}$. Subfigures: (a) $N_i = 10$, $\epsilon = 2$; (b) $M = 5$, $\epsilon =5$.}
% \label{fig:quad_sy_id}
% \end{figure*}

\subsection{Results for pendulum}
Figure~\ref{fig:pend_sy_id_SGD} demonstrates the system behavior under stochastic gradient descent (SGD). Compared to ~\ref{fig:pend_sy_id}, SGD improves the convergence rate.
\subsection{Results for quadrotor}
Another real-world system considered is the quadrotor, adapted from the model proposed in~\cite{musavi2024identification,alaimo2013mathematical}, with necessary modifications as described in the following. Let \( p \in \mathbb{R}^3 \) and \( v \in \mathbb{R}^3 \) denote the position and velocity of the quadrotor's center of mass in the inertial frame, respectively. Let \( \omega \in \mathbb{R}^3 \) represent the angular velocity in the body-fixed frame, and \( q \in \mathbb{R}^4 \) represent the orientation of the quadrotor using a unit quaternion. The corresponding equations of motion for the quadrotor are given by:
\begin{align*}
    \dfrac{d}{dt} \begin{pmatrix}
        p\\
        v\\
        q\\
        \omega
    \end{pmatrix} = \begin{pmatrix}
        v\\
        - g e_{z} + \frac{1}{m} Q f_{u}\\
        \frac{1}{2} \Omega q\\
        I^{-1} (\tau_{u} - \omega \times I \omega) 
    \end{pmatrix} + w,
\end{align*}
where \( g \) is the gravitational constant, \( m \) is the total mass of the quadrotor, \( I = \emph{diag}(I_{xx}, I_{yy}, I_{zz}) \) denotes the inertia matrix in the body-fixed frame, \( f_{u} \in \mathbb{R} \) is the total thrust, \( \tau_{u} \in \mathbb{R}^{3} \) is the total moment expressed in the body-fixed frame, and \( e_{z} = (0, 0, 1)^{\top} \) is the unit vector along the inertial \( z \)-axis.
There are seven unknown parameters in total, with heterogeneity introduced in a subset of these parameters.
\begin{align*}
    \theta_{1} &= \frac{1}{m}+\gamma_1^{(i)}, \quad
    \theta_{2} = \frac{1}{I_{xx}}, \quad
    \theta_{3} = \frac{I_{yy}-I_{zz}}{I_{xx}}, \\
    \theta_{4} &= \frac{1}{I_{yy}}, \quad
    \theta_{5} = \frac{I_{zz}-I_{xx}}{I_{yy}}, \quad
    \theta_{6} = \frac{1}{I_{zz}}, \\
    \theta_{7} &= \frac{I_{xx}-I_{zz}}{I_{zz}}.
\end{align*}
Here $\gamma_1^{(i)}\sim \mathrm{U}(0, \epsilon)$.

Figure~\ref{fig:quad_sy_id} demonstrates the experimental results for quadrotor. Figure~\ref{fig:quad_sy_id}(a) demonstrates the impact of client participation on estimation error. With a fixed trajectory count per client ($N_i = 10$) and  system heterogeneity ($\epsilon = 2$), increasing the number of participating clients  reduces the error. Figure~\ref{fig:quad_sy_id}(b) underscores the role of data volume in convergence. Keeping the number of clients fixed ($M_i = 5$) and heterogeneity ($\epsilon=5$) as fixed, increasing trajectories per client significantly accelerates error reduction. The length of each trajectory is set to 10 timesteps. These results follow similar trends as seen in the experiments for pendulum and synthetic datasets presented in the main paper.

\end{document}